%% file: main.tex
\documentclass[letterpaper]{article} 
\usepackage{aaai25}  
\usepackage{times}  
\usepackage{helvet}  
\usepackage{courier}  
\usepackage[hyphens]{url}  
\usepackage{graphicx} 
\usepackage{natbib}  
\usepackage{caption} 
\usepackage{booktabs}
\usepackage{amssymb}
\usepackage{amsmath}
\usepackage{tkz-graph}
\usepackage[procnumbered,ruled,linesnumbered,vlined ]{algorithm2e} 
\newcounter{algoline}
\newcommand{\nlast}{\refstepcounter{algoline}\nlset{\rlap{\textsuperscript{*}}}}
\SetKwInput{KwInput}{Inputs}
\SetKwInput{KwOutput}{Output}
\SetKwFor{ForEach}{for each}{do}{end}
\usepackage{pgfplots}
\pgfplotsset{compat=newest}

\usepackage{newfloat}
\usepackage{listings}
\DeclareCaptionStyle{ruled}{labelfont=normalfont,labelsep=colon,strut=off} 
\title{Resource Constrained Pathfinding with Enhanced Bidirectional {A*} Search}
\author{Saman Ahmadi\textsuperscript{*\rm 1},
Andrea Raith\textsuperscript{\rm 2},
Guido Tack\textsuperscript{\rm 3},
Mahdi Jalili\textsuperscript{\rm 1}}
\affiliations{
\textsuperscript{\rm 1} School of Engineering, RMIT University, Australia\\
\textsuperscript{\rm 2} Department of Engineering Science, University of Auckland, New Zealand\\
\textsuperscript{\rm 3} Department of Data Science and AI, Monash University, Australia\\
    *saman.ahmadi@rmit.edu.au
}

\begin{document}

\maketitle

\begin{abstract}
The classic Resource Constrained Shortest Path (RCSP) problem aims to find a cost optimal path between a pair of nodes in a network such that the resources used in the path are within a given limit.
Having been studied for over a decade, RCSP has seen recent solutions that utilize heuristic-guided search to solve the constrained problem faster.
Building upon the bidirectional A* search paradigm, this research introduces a novel constrained search framework that uses efficient pruning strategies to allow for accelerated and effective RCSP search in large-scale networks.
Results show that, compared to the state of the art, our enhanced framework can significantly reduce the constrained search time, achieving speed-ups of over to two orders of magnitude. 
\end{abstract}

\section{Introduction}
Often studied in the context of network optimization, the Resource Constrained Shortest Path (RCSP) problem seeks to find a cost-optimal path between two nodes in a network such that the resource usage of the path is within a given limit.
Its applications span a wide range of domains.
As a core problem, RCSP can be modelled to find minimum time paths for an unmanned aerial vehicle such that the total cumulative noise disturbance created by the vehicle along its trajectory is within an allowable noise disturbance level \cite{cortez2022path}.
As a subroutine, RCSP appears as a form of pricing problem in column generation (a solving technique in linear programming) for routing and scheduling applications \cite{ParmentierMV23}.
The problem has been shown to be NP-hard \cite{handler1980dual}.

RCSP is a well-studied topic in the AI and mathematical optimization literature.
A summary of conventional solutions to RCSP (methods that solve the problem to optimality) can be found in \citet{pugliese2013survey} and \citet{FeroneFFP20}. 
Among the recent solutions, the \textsf{RCBDA*} algorithm of \citet{thomas2019exact} is a label setting method that explores the network bidirectionally using heuristic-guided {A*} search.
Built on the bidirectional dynamic programming algorithm of \citet{RighiniS06}, \textsf{RCBDA*} defines an explicit perimeter for the search of each direction by evenly distributing the resource budget between the directions.
More precisely, each direction explores only those labels whose resource consumption is not more than half of the resource budget, while enabling the construction of complete paths by joining forward and backward labels.
The authors reported that the guided search of {A*} helps \textsf{RCBDA*} perform faster in large graphs when compared to the conventional RSCP methods of \citet{LozanoM13} and \citet{sedeno2015enhanced}.
Nevertheless, it still leaves several instances with a single resource constraint unsolved, even after a five-hour timeout.
\textsf{BiPulse} \cite{cabrera2020exact} is another bidirectional search framework that parallelizes the branch-and-bound \textsf{Pulse} procedure of \citet{LozanoM13}.
\textsf{BiPulse} utilizes a queueing approach to explore branches longer than a specific depth limit separately in a breadth-first search manner.
The authors reported better performance with \textsf{BiPulse} compared to \textsf{RCBDA*}, though no direct comparison was made as \textsf{RCBDA*}'s runtimes were scaled from the original paper.
Both \textsf{RCBDA*} and \textsf{BiPulse} were recognized with awards \cite{GloverPrize}.

Quite recently, the unidirectional A*-based constrained search proposed in \citet{ren2023erca}, called \textsf{ERCA*}, has demonstrated superior performance compared to \textsf{BiPulse}.
\citeauthor{ren2023erca} adapted their multi-objective search algorithm \textsf{EMOA*} \cite{RenZRLC22} to RCSP, and proposed a solution that appeared effective in reducing the constrained search effort through pruning rules enhanced with binary search trees.
Although not compared against \textsf{RCBDA*}, \textsf{ERCA*} was reported to outperform \textsf{BiPulse} by several orders of magnitude.
The two recent \textsf{WCBA*} \cite{AhmadiTHK22_socs} and \textsf{WCA*} \cite{AhmadiTHK23_Networks} algorithms can also be seen as heuristic-guided approaches to RCSP, but their application is limited to instances with single constraint only (the weight constrained variant).
Both algorithms were reported to perform faster than \textsf{RCBDA*} and \textsf{BiPulse}.

This paper introduces an enhanced bidirectional {A*} search algorithm for RCSP, called \textsf{RCEBDA*}. 
Built on the basis of perimeter search with bidirectional {A*}, our proposed algorithm leverages the recent advancements in both constrained and multi-objective search, and adopts optimized dominance pruning and path matching procedures that help reduce the search effort substantially.
Experimental results on benchmark instances reveal that \textsf{RCEBDA*} can effectively solve more instances than the state of the art, achieving speed-ups above two orders of magnitude.

\section{Problem Definition}
Consider a directed graph $G$ with a set of states $S$ and a set of edges $E \subseteq S \times S$ where every edge has $\mathit{k} \in \mathbb{N}_{\geq 2}$ attributes, represented as $(\mathit{cost}, \mathit{resource}_1, \dots, \mathit{resource_{k-1}})$.
A path on $G$ is a sequence of states $u_i \in S$ with $i \in \{1, \dots, n \}$ and $(u_i,u_{i+1}) \in E$ for $i \in \{1,\dots, n-1\}$.
The cost and resource consumption of path $\pi=\{u_1,\dots,u_n\}$ is then the sum of corresponding attributes on all the edges along the path.
The objective in the point-to-point RCSP problem is to find a $\mathit{cost}$-optimal path $\pi^*$ between a given pair of ($\mathit{start},\mathit{goal}$) states such that the consumption of resources on the path respects a set of limits ${\bf R}=\{R_1,\dots,R_{k-1}\}$, that is, we must have $\mathit{resource_{i}}(\pi^*) \leq R_i$ for every $i \in \{1,\dots,k-1\}$.

Our notation generalizes the two possible $\textit{forward}$ and $\textit{backward}$ search directions by searching in direction~$d$ from an $\textit{initial}$ state to a $\textit{target}$ state.
A forward search explores the forward graph, denoted by $G^f=(S,E^f)$, from the $\mathit{start}$ state, while a backward search begins from the $\mathit{goal}$ state on the reversed graph (i.e., $G^f$ with all its edges reversed), denoted by $G^b=(S,E^b)$.
Further, $\mathit{Succ}^d(u)$ denotes all successor states of $u$ in the graph of search direction $d$.

To simplify our algorithmic description, we define edge attributes as a form of cost vector, namely $\mathit{\bf cost} = (\mathit{cost}_1, \dots, \mathit{cost_{k}})$, with $\mathit{cost}_1$ denoting the primary cost of the problem and ($\mathit{cost}_2, \dots, \mathit{cost_{k}})$ representing the resources.
Edge costs in the search graph of direction $d$ can be accessed through the function ${\bf cost}^d:E^d \to \mathbb{R}_{\geq 0}^{k}$.
The search objects in our notation are defined as \textit{nodes}, which can be seen as partial paths obtained through exploring the graph from the $\mathit{initial}$ state.
A node $x$ conventionally contains some key information on the partial path extended from the $\mathit{initial}$ state to state $s(x) \in S$.
The function $s(x)$ here returns the state associated with $x$. 
Node $x$ traditionally stores $\mathit{\bf cost}$ of the partial path, accessible via the value pair $\mathbf{g}(x)$.
It also stores a value pair $\mathbf{f}(x)$, which estimates the $\mathit{\bf cost}$ of a complete $\mathit{initial}$-$\mathit{target}$ path via $x$; and also a reference $\mathit{parent}(x)$ which indicates the parent node of $x$.

All operations on the vectors (shown in boldface) are assumed to be done element-wise.
We use the symbol $\preceq$ in direct comparisons of cost vectors.
To compare vectors for their resource component only, we use the symbol $\preceq^\mathrm{Tr}$ to denote the truncation of the first cost in the comparison, e.g., ${\bf g}(x) \preceq^\mathrm{Tr} {\bf g}(y)$ denotes
${g_i}(x) \leq {g_i}(y)$ for all $i \in \{2,\dots,k\}$.

\noindent \textbf{Definition\ }
Let $\overline{\bf f} = (\overline{f_1},\dots,\overline{f_k})$ be a global upper bound on $\mathit{\bf cost}$ of any $\mathit{start}$-$\mathit{goal}$ paths.
$\mathit{x}$ is out of bounds if ${\bf f}(x) \npreceq \overline{{\bf f}}$. 

\noindent \textbf{Definition\ }
Node $\mathit{x}$ weakly dominates node $\mathit{y}$ if we have ${\bf g}(x) \preceq {\bf g}(y)$; $\mathit{x}$ (strongly) dominates $\mathit{y}$ if ${\bf g}(x) \preceq {\bf g}(y)$ and ${\bf g}(x) \neq {\bf g}(y)$; $y$ is not dominated by $x$ if ${\bf g}(x) \npreceq {\bf g}(y)$.

In the context of bidirectional {A*}, the search in direction $d$ is guided by cost estimates or {\bf f}-values, which are traditionally established based on a heuristic function
${\bf h}^d: S \rightarrow \mathbb{R}_{\geq 0}^{k}$ 
\cite{hart1968formal}.
In other words, for every search node $x$ in direction $d$, we have ${\bf f}(x)={\bf g}(x)+{\bf h}^d(s(x))$ where ${\bf h}^d(s(x))$ returns lower bounds on the ${\bf cost}$ of paths from $s(x)$ to the $\mathit{target}$ state.

\noindent \textbf{Definition} 
The function ${\bf h}^d: S \rightarrow \mathbb{R}^k_{\geq0}$ is consistent if ${\bf h}^d(u) \le {\bf cost}^d(u,v) + {\bf h}^d(v)$ for every edge $(u,v) \in E^d$. 
It is also admissible if ${\bf h}^d(\mathit{initial}) = {\bf 0}$.

\section{Bidirectional Constrained {A*} Search}
Originally introduced by \citet{pohl1971bi} for single-objective pathfinding, bidirectional search strategies have since evolved, with newer approaches offering promising advancements in search efficiency \cite{Alcazar21,SiagSFS23}. 
In the RCSP context, the classic front-to-end bidirectional A* search of \textsf{RCBDA*} has proven effective in guiding constrained searches to optimal solutions \cite{thomas2019exact,AhmadiTHK21}. 
Analogous to classic bidirectional {A*} search, the \textsf{RCBDA*} algorithm employs an interleaved strategy in which the search is guided by the direction that exhibits the lowest ${f}_1$-value.
Let $f^*_1$ be the optimal cost of a valid solution path.
\textsf{RCBDA*} enumerates in best-first order all partial paths with $f_1$-value no larger than $f^*_1$ in both directions sequentially.
A complete $\mathit{start}$-$\mathit{goal}$ path in this framework is built by joining forward and backward partial paths arriving to the same state.
However, since the number of paths to each state can grow exponentially during the search, partial path matching can become a demanding task if each direction is allowed to explore the entire search space.

To mitigate the bottleneck above, \textsf{RCBDA*} utilizes the perimeter bounding strategy proposed by \citet{RighiniS06}, and equally distributes one of the resource budgets between the directions.
The selected resource is referred to as the \textit{critical} resource.
Consequently, the algorithm explores, in each direction, those partial paths that consume no more than half of the provided critical resource budget.
In addition, to prevent the search from exploring unpromising paths, \textsf{RCBDA*} does not process nodes that have already exhausted (any of) the resource budgets.
\citeauthor{thomas2019exact} experimented \textsf{RCBDA*} with the pruning rules presented in \citet{LozanoM13}, namely: pruning by dominance and infeasibility.
The former allows the search to prune nodes dominated by a previously explored node, while the latter ensures that nodes with estimated resource usage larger than the given limits are not explored.
Although the authors did not discuss the pruning rules as part of the proposed label-setting approach, we have provided in Algorithm~\ref{alg:rcbda} a refined pseudocode closest to their description, including both pruning rules.

\input{Algs/RCBDA}
\input{Algs/IsDom}

The algorithm starts with initializing a global search upper bound $\overline{\bf f}$ using the resource limits provided.
It then establishes the heuristic functions of the forward and backward direction ${\bf h}^f$ and ${\bf h^b}$, respectively.
These heuristic functions are conventionally obtained through $k$ rounds of single-objective one-to-all searches (e.g., using the Dijkstra's algorithm \cite{johnson1973note}) in the reverse direction.
The algorithm also initializes for each direction $d$ two types of data structures: a priority queue, called $\mathit{Open}^d$, that maintains unexplored nodes generated in direction $d$; and a node list $\mathcal{X}^d(u)$ that stores nodes explored with state $u \in S$ in direction $d$.
$\mathcal{X}^d(u)$ is used to fulfil the pruning by dominance rule, as well as path matching with nodes explored in the opposite direction.
To commence the search, the algorithm generates two initial nodes, one for each direction, and inserts them into the corresponding priority queue.

The main constrained search starts at line~\ref{alg:rcbda:iter} and continues until both priority queues are empty.
Let $\overline{f_1}$ be the best known upper bound on the primary cost of the problem (initially unknown).
In each iteration, the algorithm extracts from one of the queues a node $x$ that has the smallest $f_1$-value among all unexplored nodes.
If there are multiple candidates, it breaks ties arbitrarily.
The search can stop immediately if we find $f_1(x) \geq \overline{f_1}$, essentially because it has proven the optimality of $\overline{f_1}$.
Note that {A*} explores nodes in non-decreasing order of their $f_1$-value. Thus, all unexplored nodes in the priority queues would not improve $\overline{f_1}$ upon fulfillment of the termination criterion of line~\ref{alg:rcbda:terminate}.

Let $d$ be the direction of the priority queue from which $x$ was extracted.
Also assume $d'$ is the opposite direction of $d$.
If the search is not terminated, the algorithm expands $x$ if the consumption of its critical resource, denoted by $g_\kappa(x)$ for $ \kappa \in \{2,\dots,k\}$, is not more than 50\% of the total budget, or $\overline{f_\kappa}/2$ equivalently (line~\ref{alg:rcbda:half}). 
Expansion of $x$ generates a set of descendant nodes.
Each descendant node $y$ will be added into $\mathit{Open}^d$ unless it is deemed out of bounds (line~\ref{alg:rcbda:bound}), a process known as pruning by infeasibility.
The algorithm then checks the descendant node against previously explored nodes (known as pruning by dominance).
As scripted in Algorithm~\ref{alg:Isdominated}, the dominance check involves comparing ${\bf g}(y)$ against $\mathit{\bf cost}$ of all previously explored nodes with $s(y)$ in $\mathcal{X}^d(s(y))$. 
Since nodes with $s(y)$ are explored in non-decreasing order of their $g_1$-values, we optimize the operation and compare nodes for their resource component only (using the $\preceq^{\mathrm{Tr}}$ operator).

The next step involves constructing complete $\mathit{start}$-$\mathit{goal}$ paths by joining $x$ with nodes explored with $s(x)$ in the opposite direction $d'$, already available in $\mathcal{X}^{d'}(s(x))$.
The search updates the best known upper bound on the primary cost, i.e., $\overline{f_1}$, if it finds that the joined path is within the bounds (lines~\ref{alg:rcbda:match1}-\ref{alg:rcbda:match2}).
After the completion of the path matching step, $x$ will be added to the list of expanded nodes with $s(x)$ in direction $d$ (line~\ref{alg:rcbda:add}), to be joined with nodes explored in the opposite direction.
Finally, the algorithm returns $\overline{f_1}$ as an optimal cost to the given RCSP instance.

\section{An Enhanced Bidirectional RCSP Search}
As explained in the previous section, \textsf{RCBDA*} stores in both directions all explored nodes of each state so it can prune dominated nodes before inserting them into the priority queue.
Although this approach helps alleviate the queue load by removing dominated nodes as soon as they are generated, it may still let the search explore some dominated nodes.
This is because the algorithm does not check newly generated nodes against unexplored nodes, which can lead to having dominated nodes in the queues.
Nonetheless, given that the number of explored nodes grows exponentially during the search, it is likely that \textsf{RCBDA*} spends most of its search time on dominance pruning.
This costly pruning procedure can be considered a search bottleneck, yet it is essential to prevent the frontier from growing uncontrollably.

Besides dominance pruning, search initialization can be seen as a crucial task in improving the efficiency of the main search.
\textsf{RCBDA*} is conventionally initialized using multiple rounds of one-to-all searches, which yield the {A*}'s required heuristics.
Although the correctness of constrained {A*} search relies on its heuristic being consistent and admissible, it has been shown that the search performance can be greatly impacted with the quality of heuristics \cite{AhmadiTHK23_Networks}, suggesting further opportunities for enhancing the exhaustive resource constrained search with {A*}.

To address the above shortcomings, this research presents \textsf{RCEBDA*}, a new RCSP algorithm that enhances the resource constrained search with bidirectional {A*}.
A pseudocode of \textsf{RCEBDA*} is provided in Algorithm~\ref{alg:rcebda}, with the symbol $*$ next to specific line numbers indicating our proposed changes.
We describe each enhancement as follows.

\input{Algs/RCEBDA-Init}
\textbf{Better informed heuristics:}
When conducting {A*} bidirectionally, we need to compute both backward and forward lower bounds.
In the context of RCSP, this lower bounding strategy provides us with a great opportunity to feed the main search with better informed heuristics, thereby reducing the search effort and node expansions, as shown by \citet{AhmadiTHK21,AhmadiTHK23_Networks} for the weight constrained variant.
We present in Algorithm~\ref{alg:initialization} our proposed initialization phase for \textsf{RCEBDA*}.
After initializing the lower bounds, the algorithm runs $k$ rounds of one-to-all searches in both directions, starting from the last attribute.
In each round, we apply a technique called \textit{resource-based network reduction} \cite{AnejaAN83} and use the established lower bounds to prune out-of-bounds states.
Given $\overline{f_i}$ as the global upper bound on $\mathit{cost_i}$ of paths, state $u\in S$ can be removed if its least $\mathit{cost}_i$ paths to both ends yield a complete path that violates $\overline{f_i}$ (line~\ref{alg:init:prune}).
This pruning can be skipped in the last round ($i=1$), as $\overline{f_1}$ is initially unknown.
Consequently, there will be $k-1$ levels of graph reduction, all contributing to improving the quality of the heuristics.

\input{Algs/RCEBDA}
\textbf{Efficient dominance pruning:}
The fact that nodes in {A*} are explored in non-decreasing order of their $f_1$-value enables us to reduce the dimension of cost vectors by one in all dominance checks.
In other words, a node $x$ is weakly dominated by a previously explored node $y$, $s(x)=s(y)$, if ${\bf g}(y) \preceq^\mathrm{Tr} {\bf g}(x)$.
Otherwise, $x$ will be stored in $\mathcal{X}^d(s(x))$ if it is deemed non-dominated.
In the latter case, if we observe ${\bf g}(x) \preceq^\mathrm{Tr} {\bf g}(y)$, we can guarantee that any future node with $s(x)$ and weakly dominated by $y$ will similarly be weakly dominated by $x$, rendering the storage of $y$ in $\mathcal{X}^d(s(x))$ redundant.
Previously studied in \citet{NAMOA15}, this technique has been utilized in both constrained and multi-objective search to reduce the dominance check effort by removing dominated (truncated) cost vectors from the expanded list \cite{ren2023erca,LTOMOA3}.
In our bidirectional framework, however, such nodes cannot be fully removed from the search, as every non-dominated node must be available for potential path matching.
To this end, we allow each state $u \in S$ to store its previously explored nodes in two lists: $\mathcal{X}^d(u)$ and $\mathcal{X}^d_{{Dom}}(u)$.
The latter maintains nodes whose truncated cost vector is dominated by that of one node in $\mathcal{X}^d(u)$, while the former stores non-dominated ones.
Note that neither $\mathcal{X}^d(u)$ nor $\mathcal{X}^d_{{Dom}}(u)$ contains weakly dominated nodes.
In addition, to ensure nodes are checked for dominance before expansion, we employ a lazy approach and delay the dominance check of nodes until they are extracted from the queue (line~\ref{alg:rcebda:dom}).
Although this approach incurs extra queueing effort due to queues containing more dominated nodes, it will lead to reducing node expansions. 

Let $x$ be a non-dominated node just extracted from $\mathit{Open}^d$.
This node will eventually be added to $\mathcal{X}^d(s(x))$, but to keep the truncated cost vector of nodes in $\mathcal{X}^d(s(x))$ non-dominated, we first check other nodes of the list against $x$.
As scripted in lines~\ref{alg:rcebda:removedom1}-\ref{alg:rcebda:removedom3} of Algorithm~\ref{alg:rcebda}, all nodes whose truncated cost vector is dominated by that of $x$ are removed from $\mathcal{X}^d(s(x))$ and added into $\mathcal{X}^d_{Dom}(s(x))$ subsequently.

\textbf{Quick dominance pruning:}
We also utilize a quick pruning rule that checks $x$ against the most recent node explored with $s(x)$ before the rigorous dominance check, and also during expansion for descendants of $x$ (see lines~\ref{alg:rcebda:quick1}-\ref{alg:rcebda:quick2} and \ref{alg:rcebda:quick3}-\ref{alg:rcebda:quick4}).
This technique has proven to be effective in substantially reducing the number of dominance checks during exhaustive multi-objective A* search \cite{AhmadiSHJ24}.

\input{Algs/Match}
\textbf{Path matching:}
\textsf{RCEBDA*} stores all explored but non-dominated nodes of each state in two lists, so it can still join bidirectional paths.
Let $x$ be a non-dominated node explored in direction $d$. 
Scripted in Algorithm~\ref{alg:match}, \textsf{RCEBDA*} joins $x$ with all non-dominated nodes of the opposite direction $d'$, stored in $\mathcal{X}^{d'}(s(x))$ and $\mathcal{X}^{d'}_{Dom}(s(x))$.
Contrary to \textsf{RCBDA*} where it captures the optimal cost only, we allow the algorithm to store all node pairs yielding a non-dominated $\mathit{cost}_1$-optimal solution path.
For every candidate node $y$ of opposite direction, the algorithm checks whether joining $x$ with $y$ constructs a feasible path (line~\ref{alg:match:bound}). 
If the joined path improves the best known optimal cost $\overline{f_1}$, we update the global upper bound and remove all previous (suboptimal) solutions (line~\ref{alg:match:update1}-\ref{alg:match:update2}).
The algorithm then explicitly checks the joined path against all existing solutions for dominance (lines~\ref{alg:match:dom1}-\ref{alg:match:dom2}).
The node pair representing the joined path will be added to $\mathit{Sols}$ if the algorithm finds the joined path non-dominated (line~\ref{alg:match:nondom}).
To keep the solution set only containing nodes that form non-dominated solution paths, the algorithm removes from $\mathit{Sols}$ node pairs weakly dominated by the joined path.
Note that the algorithm allows successive updates to the upper bound when a better solution is found, and all node pairs in $\mathit{Sols}$ always have the same $f_1$-value.

Although the path matching procedure of Algorithm~\ref{alg:match} considers all non-dominated nodes explored in the opposite direction, we can optimize the procedure by prioritizing the nodes in the $\mathcal{X}^{d'}(s(x))$ list.
In this optimized strategy, if we find all joined paths out of bounds for their resource usage, that is, 
if we have ${\bf g}(x) + {\bf g}(y) \npreceq^\mathrm{Tr} \overline{\bf f}$ for all $y \in \mathcal{X}^{d'}(s(x))$, we can skip matching $x$ with candidates from $\mathcal{X}^{d'}_{Dom}(s(x))$ because they are already dominated by a node in $\mathcal{X}^{d'}(s(x))$ for their resources, rendering their matching unnecessary.

A formal proof of the correctness of \textsf{RCEBDA*} is provided in 
the next section.
The example below further elaborates on the key steps involved in the algorithm.

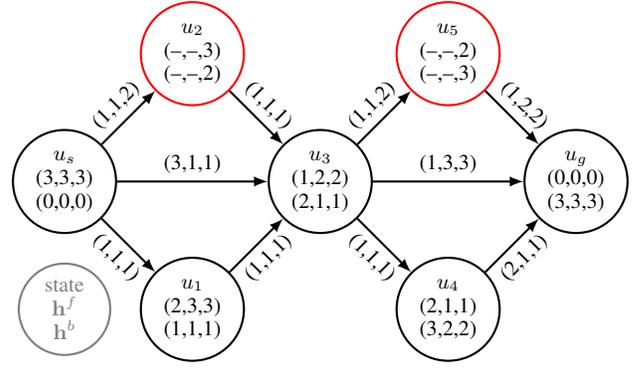
\begin{figure}
\input{Figures/Example}
\caption{\small An example graph with $k=3$ and $R=\{4,4\}$. Triples inside the states denote ${\bf h}^d$. $u_2$ and $u_5$ can be seen out of bounds.}
\label{fig:example}
\end{figure}

\textbf{Example:}
Consider the sample RCSP instance of Figure~\ref{fig:example} with three edge attributes (two resources) with resource limits $R=\{4,4\}$.
We aim to find an optimal feasible path from state $u_s$ to state $u_g$.
\textsf{RCEBDA*} first sets the upper bound $\overline{\bf f} \gets (\infty,4,4)$.
For every state, we have provided the forward and backward lower bounds as triples within the states.
Note that each $h^d_i$-value represents a $\mathit{cost}_i$-optimal path for $i \in \{1,2,3\}$.
Checking the states for their bidirectional lower bounds, we realize that states $u_2$ and $u_5$ (shown in red) are out of bounds for their last cost, as we have $3+2 \nleq 4$.
Thus, they can be removed from the search graph in the initialization phase.
Let the last (third) attribute be the critical resource for the perimeter search of \textsf{RCEBDA*}, i.e., $\kappa = 3$.
Thus, each direction is given half of the critical resource budget, or equivalently $\overline{f_\kappa}/2 = 2$, meaning that nodes with $g_3$-value greater than 2 are not expanded.
We now briefly explain iterations (It.) of \textsf{RCEBDA*} that lead to finding the first solution path.
The iterations are shown in Table~\ref{tab:example}.
For each direction $d$, we show the trace of its priority queue $\mathit{Open}^d$, and also, in separate columns, changes on $\mathcal{X}^d$ and $\mathcal{X}^d_{Dom}$ of expanded states during the iteration.
Expanded nodes are denoted by an asterisk next to them ($^*$).
The last column shows changes on $\mathit{Sols}$.
Before the first iteration starts, the algorithm inserts one initial node to the priority queue of each direction.
If both queues exhibit the same smallest $f_1$-value, we prioritize the forward search.

\input{Tables/example}
\noindent
\textbf{It.1:} 
Node $x_1$, with $u_s$, is extracted from the forward direction.
$x_1$ is not dominated, thus it will be stored in $\mathcal{X}^f(u_s)$.
Its critical resource usage is also within the predefined perimeter.
Thus, $x_1$ will be expanded, yielding two new nodes $x_3$ and $x_4$ with states $u_1$ and $u_3$, respectively.
Note that $u_2$ is already removed from the search graph.
Both nodes will be added to $\mathit{Open}^f$ since they are within the bounds.

\noindent
\textbf{It.2:} $x_3$ is extracted (with $u_1$).
This node is non-dominated, and thus can be stored in $\mathcal{X}^f(u_1)$.
Since $x_3$'s critical resource usage is smaller than the allocated budget, it will be expanded. Thus, a new node $x_5$ will be added to the queue.

\noindent
\textbf{It.3:}
$x_5$ is extracted (with $u_3$).
$x_5$ is non-dominated and can be expanded.
Two new nodes $x_6$ (with $u_g$) and $x_7$ (with $u_4)$ are generated.
The first node $x_6$ appears out of bounds as we obtain ${\bf f}(x_6) = (3,5,5)$ during the expansion.
Node $x_7$, however, is within the bounds and will be added to $\mathit{Open}^f$.

\noindent
\textbf{It.4:}
Now $x_2$ is extracted from the \textit{backward} direction.
Expansion of $x_2$ adds two new nodes $x_8$ and $x_9$ into $\mathit{Open}^b$.

\noindent
\textbf{It.5:}
$x_8$ from the backward direction is extracted.
$x_8$ is non-dominated and thus will be stored in $\mathcal{X}^b(u_3)$.
However, $x_8$ is not expanded as its critical resource usage is not within the allocated budget, i.e, we have $g_3(x_8) \nleq 2$.

\noindent
\textbf{It.6:}
$x_4$ from the \textit{forward} direction is extracted.
There is already one node, $x_5$, in the $\mathcal{X}^f(u_3)$ list, which cannot dominate $x_4$.
However, we observe that the truncated cost vector of $x_5$ is dominated by that of $x_4$. 
Thus, $x_5$ will be moved to $\mathcal{X}_{Dom}^f(u_3)$ before adding $x_4$ to $\mathcal{X}^f(u_3)$.
Checking the nodes explored with $u_3$ in the backward direction, we find $x_8$ as a candidate for path matching.
Joining $x_4$ with $x_8$ yields a complete path that is within the bounds, i.e., ${\bf g}(x_4) + {\bf g}(x_8) = (4,4,4) \preceq \overline{\bf f}$.
Consequently, the upper bound on primary cost can be updated, and $(x_4,x_8)$ can be captured as a solution node pair.

\input{Appendix}

\section{Empirical Analysis}
This section evaluates the performance of our proposed algorithm and compares it against the state of the art.
For the baseline, we consider the recent \textsf{ERCA*} algorithm \cite{ren2023erca} and the improved version of \textsf{RCBDA*} \cite{thomas2019exact} we presented in Algorithm~\ref{alg:rcbda}.

\textbf{Benchmark:}
We use the instances of \citet{AhmadiTHK23_Networks}, which are defined over the road networks from the 9th DIMACS Implementation Challenge \cite{dimacs9th}.
We choose 20 ($\mathit{start}$,$\mathit{goal}$) pairs over three maps, with the largest map containing over 400K nodes and 1M edges.
We study two RCSP scenarios: $k=3$ (two resources) and $k=4$ (three resources). 
The first and second edge costs are \textit{distance} and \textit{time}, respectively.
Following \citet{ren2023erca}, we choose the third cost to be the average (out)degree of the link, that is, the number of adjacent vertices of each end point.
We choose the fourth cost of each edge to be one.
Each ($\mathit{start}$,$\mathit{goal}$) pair is evaluated on five tightness levels, yielding 100 instances per map per scenario. 
Following the literature, we define for each resource $i \in \{1,\dots,k-1\}$ the budget $\mathit{R_i}$ based on the tightness of the constraint $\delta$ as:
\begin{equation*}
\delta=\dfrac{R_{i}-\mathit{cost}^{\mathit{min}}_{i+1}}{\mathit{cost}^{\mathit{max}}_{i+1}-\mathit{cost}^{\mathit{min}}_{i+1}} \quad \text{for} \ \delta \in\{10\%,30\%,\dots,90\%\} 
\end{equation*}
where $\mathit{cost}^{\mathit{min}}_{i+1}$ and $\mathit{cost}^{\mathit{max}}_{i+1}$ are lower and upper bounds on $\mathit{cost}_{i+1}$ of $\mathit{start}$-$\mathit{goal}$ paths, respectively.
Each $\mathit{cost}^{\mathit{max}}_{i+1}$ is calculated using the non-constrained $\mathit{cost}_1$-optimal path.

\textbf{Implementation:}
We implemented our \textsf{RCEBDA*} and the improved version of \textsf{RCBDA*} in C++ and used the publicly available C++ implementations of \textsf{ERCA*}.
All C++ code was compiled using the GCC7.5 compiler with O3 optimization settings. 
In addition, we implemented a variant of our \textsf{RCEBDA*} where the bidirectional searches are conducted concurrently, denoted \textsf{RCEBDA*\textsubscript{par}}.
Similar to the extended parallel weight constrained search in \citet{AhmadiTHK23_Networks}, we allocate one thread to each direction and allow the forward (resp. backward) search to work on $\mathit{Open}^f$ (resp. $\mathit{Open}^b$) independently. 
The overall search in this framework terminates when both concurrent searches are exited (see \citet{AhmadiTHK23_Networks} for more details).
We define the critical resource to be the last attribute in both \textsf{RCBDA*} and \textsf{RCEBDA*}.
In addition, following \citet{AhmadiSHJ24}, we store non-dominated nodes of each state in a dynamic array, in which nodes are maintained in lexicographical order of their truncated cost vector. 
We ran all experiments on a single core of an Intel Xeon Platinum 8488C processor running at 2.4~GHz, with 32~GB of RAM and a one-hour timeout.
Our code is publicly available\footnote{https://bitbucket.org/s-ahmadi/multiobj}.

\input{Tables/Main_results}
Table~\ref{table:results} presents the results for both scenarios.
We report the number of solved cases $|\mathcal{S}|$ and runtime statistics (in seconds, including the initialization time).
We also design a virtual best oracle, which, for each instance, provides the best runtime achieved by either of the algorithms.
The last column of the table shows the average slowdown factor $\phi$ of each algorithm compared to the virtual best oracle over mutually solved instances of each map.
We observe that, while \textsf{RCEBDA*} and \textsf{RCEBDA*\textsubscript{par}} successfully solved 100\% of instances within 20 minutes, \textsf{RCBDA*} and \textsf{ERCA*} were unable to solve several instances of each map in both scenarios within the one-hour timeout.
Interestingly, the (improved) \textsf{RCBDA*} algorithm has consistently performed better than \textsf{ERCA*} in all maps in terms of solved cases and average runtime.
Comparing the average slowdown factors, we find both \textsf{RCEBDA*} and \textsf{RCEBDA*\textsubscript{par}} performing close to the virtual best oracle, with the parallel variant showing slightly better performance.
The other two algorithms, \textsf{RCBDA*} and \textsf{ERCA*}, perform up to two orders of magnitude slower than the virtual best oracle on average, highlighting the success of our proposed algorithm in solving challenging RCSP instances faster.
Our detailed results show that the speedup is primarily driven by the proposed search engine enhancements, especially efficient dominance pruning, followed by the optimized two-stage path-matching strategy and, to a lesser extent, quick dominance pruning.
To better compare the algorithm for their performance, Figure~\ref{fig:scatter_plot} displays the runtimes of \textsf{RCEBDA*} and \textsf{RCEBDA*\textsubscript{par}} versus \textsf{RCBDA*} across all instances with three resources.
Despite the benchmark including numerous easy instances, we can observe that \textsf{RCEBDA*} and \textsf{RCEBDA*\textsubscript{par}} demonstrate superior performance in the more challenging instances, outperforming \textsf{RCBDA*} by over two orders of magnitude.
Note that the comparison of the three algorithms here is head-to-head, as they are implemented within the same framework.

\input{Figures/ScatterPlot2}
%
\begin{figure}[!t]
\centering
\includegraphics[width=0.95\columnwidth]{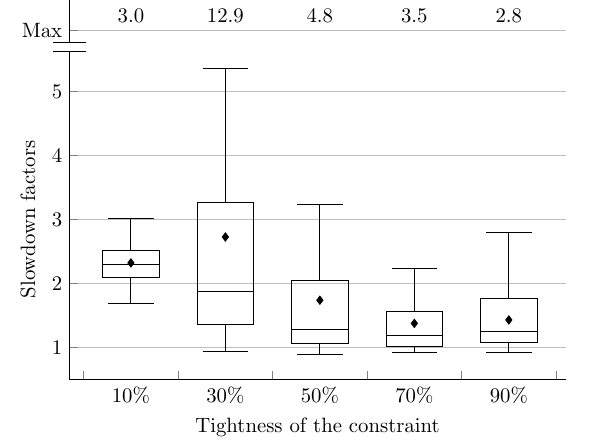}
\caption{Distribution of slowdown factors for \textsf{RCEBDA*} initialized conventionally.
Each plot shows first quartile (25\% data), median, third quartile (75\% of data), mean ({\tiny\(\blacklozenge\)}), minimum and maximum within the 1.5$\times$ interquartile range, and absolute maximum slowdown. Outliers are not shown.
}
\label{fig:cactus_performance}
\end{figure}

To study the impact of initialization, we experimented a variant of \textsf{RCEBDA*} in which the upper bound pruning of states is disabled (line~\ref{alg:init:prune} of Algorithm~\ref{alg:initialization}).
Figure \ref{fig:cactus_performance} depicts, for each tightness level, the distribution of slowdown factors obtained with this variant with respect to the standard variant across all instances with $k=4$.
We observe that, while disabling state upper bounding during initialization results in slower performance across all tightness levels, the impact is more pronounced in tight constraints (10\% and 30\%), where it leads to an average slowdown exceeding a factor of two.

\section{Conclusion}
This research presented \textsf{RCEBDA*}, an enhanced bidirectional A* search algorithm for RCSP.
Built on the half-way bidirectional search scheme in the literature, our proposed method introduces several enhancements to the bidirectional constrained A* search, including more efficient initialization, dominance pruning and partial path matching strategies.
The results of our experiments over benchmark instances from the literature demonstrate the success of \textsf{RCEBDA*} in substantially reducing the computation time of RCSP in large-scale graphs, achieving speed-ups of several orders of magnitude compared to the state of the art.

\section{Acknowledgement}
This research was supported by the Department of Climate Change, Energy, the Environment and Water under the International Clean Innovation Researcher Networks (ICIRN) program grant number ICIRN000077.
Mahdi Jalili is supported by Australian Research Council through projects DP240100963, DP240100830, LP230100439 and IM240100042.

\bibliography{References.bib}

\end{document}

%% file: Algs/RCBDA.tex
\begin{algorithm}[!t]
\small
\caption{\textsf{RCBDA*} (Improved)}
\label{alg:rcbda}
\DontPrintSemicolon
\KwIn{An RCSP Problem ($\mathit{G}, \mathit{start}, \mathit{goal},R_1,\dots,R_{k-1}$)}
 
 \KwOutput{Optimal cost of the problem}

$ \overline{\bf f} \gets (\infty, R_1, \dots, R_{k-1})$ \;

  ${\bf h}^f,{\bf h}^b \gets$ forward and backward $\mathit{\bf cost}$ lower bounds \;
  
 $\mathit{Open}^f \gets \emptyset, \mathit{Open}^b \gets \emptyset $ \;
$\mathcal{X}^f(u) \gets \emptyset, \mathcal{X}^b(u) \gets \emptyset$ $\forall u \in S$ \;
 $x \gets $ new node with $s(x) = \mathit{start}$\ \;
  $y \gets $ new node with $s(y) = \mathit{goal}$\ \;
 $ {\bf g}(x) \gets {\bf g}(y) \gets {\bf 0}$,
 $ {\bf f}(x) \gets {\bf h}^f(\mathit{start})$,
 $ {\bf f}(y) \gets {\bf h}^b(\mathit{goal})$\;

add $x$ to $\mathit{Open}^f$ and $y$ to $\mathit{Open}^b$\;

\While{$\mathit{Open}^f \bigcup \mathit{Open}^b \neq \emptyset$ \label{alg:rcbda:iter}}
{

  {extract from $\mathit{Open}^f \bigcup \mathit{Open}^b$ node $x$ with the smallest $f_1$-value \label{alg:rcbda:least_cost} \;}
       \lIf{ $f_1(x) \geq \overline{f_1}$ \label{alg:rcbda:terminate}} 
      {\textbf{break} }
      
       $d \gets$ the direction from which $x$ was extracted \;
       $d' \gets$ the opposite direction of $d$ \;
       


\If{ $g_{\kappa}(x) \leq \overline{f_{\kappa}}/2$\label{alg:rcbda:half}} 
    {
    \ForEach{$t \in \mathit{Succ}^d(s(x))$}
        {  $y \gets $ new node with $s(y) = t$ \label{alg:rcbda:expansion1}\; 
             ${\bf g}(y) \gets {\bf g}(x) + {\bf cost}^d (s(x),t)$ \; 
             ${\bf f}(y) \gets {\bf g}(y) + {\bf h}^d (t)$ \;
             
             \lIf{ ${\bf f}(y) \npreceq \overline{\bf f}$
       \label{alg:rcbda:bound}} 
      {\textbf{continue} }
      \lIf{$\mathtt{IsDominated}(y,\mathcal{X}^d(t))$ \label{alg:rcbda:dom}} 
      {\textbf{continue}}

            Add $y$ to $\mathit{Open}^d$\; \label{alg:rcbda:expansion3}
            
        }
    }
     \ForEach{$y \in \mathcal{X}^{d'}(s(x))$\label{alg:rcbda:match1}}
     {
     \lIf {${\bf g}(x) + {\bf g}(y) \preceq {\overline{\bf f}}$ }
        {$\overline{f_1} \gets g_1(x) + g_1(y)$ \label{alg:rcbda:match2}}
    
     }

     add $x$ to $\mathcal{X}^d(s(x))$\label{alg:rcbda:add}\;
}
\Return{$\overline{f_1}$}
\end{algorithm}

%% file: Algs/IsDom.tex
\begin{algorithm}[!ht]
\small
\caption{$\mathtt{IsDominated}$}
\label{alg:Isdominated}
\DontPrintSemicolon
\KwIn{A node $x$ and a set of nodes $\chi$}
 
 \KwOutput{$\mathit{true}$ if $x$ is weakly dominated, $\mathit{false}$ otherwise}

    \ForEach{$y \in \chi$}
        {  
        
        \lIf {${\bf g}(y) \preceq^\mathrm{Tr} {\bf g}(x)$ \label{alg:isdom:check1}}
        {\Return{$\mathit{true}$}}
        
        }
\Return{$\mathit{false}$}
\end{algorithm}

%% file: Algs/RCEBDA-Init.tex
\begin{algorithm}[t]
\small
\caption{$\mathtt{Initialize}$}
\label{alg:initialization}
\DontPrintSemicolon
\KwIn{An RCSP instance ($\mathit{G}$, $\mathit{\mathit{start}}$, $\mathit{\mathit{goal}}, \overline{\bf 
 f}$)}

\KwOutput{Lower bounds $({\bf h}^f,{\bf h}^b)$, $S$ updated }

$ {\bf h}^f(u) \gets {\bf h}^b(u) \gets {\bf \infty}^{k} \ \forall u \in S$ \;
    \For{$ i \in \{k, \dots, 1 \}$}
        {  
        $h^f_i \gets$ Single-objective backward search on $\mathit{cost}_i$\;
        $h^b_i \gets$ Single-objective forward search on $\mathit{cost}_i$\;
        $S\gets \{ u \ | \ u \in S \ \mathrm{and} \ h^f_i(u) + h^b_i(u) \leq \overline{f_i} \}$\label{alg:init:prune}\;
        }


 

  
\Return{ $({\bf h}^f, {\bf h}^b)$}
\end{algorithm}

%% file: Algs/RCEBDA.tex
\begin{algorithm}[!t]
\small
\caption{\textsf{RCEBDA*} }
\label{alg:rcebda}
\DontPrintSemicolon
\KwIn{An RCSP problem ($\mathit{G}$, $\mathit{\mathit{start}}$, $\mathit{\mathit{goal}}, R_1, \dots, R_{k-1}$)}

\KwOutput{A set of node pairs representing optimal paths}

$ \overline{\bf f} \gets (\infty, R_1, \dots, R_{k-1})$ \;
\nlast
${\bf h}^f, {\bf h}^b \gets \mathtt{Initialize}$($\mathit{G}$, $\mathit{\mathit{start}}$, $\mathit{\mathit{goal}}, \overline{\bf 
 f}$)\;
\nlast
$\mathit{Sols} \gets \emptyset $\;
$\mathit{Open}^f \gets \emptyset, \mathit{Open}^b \gets \emptyset $ \;
$\mathcal{X}^f(u) \gets \emptyset, \mathcal{X}^b(u) \gets \emptyset$ $\forall u \in S$ \;
\nlast
$\mathcal{X}_\mathit{Dom}^f(u) \gets \emptyset, \mathcal{X}_\mathit{Dom}^b(u) \gets \emptyset$ $\forall u \in S$ \;
 $x \gets $ new nodes with $s(x) = \mathit{start}$\ \;
$y \gets $ new node with $s(y) = \mathit{goal}$\ \;
 $ {\bf g}(x) \gets {\bf g}(y) \gets {\bf 0}$ ,
 $ {\bf f}(x) \gets {\bf h}^f(\mathit{start})$,
 $ {\bf f}(y) \gets {\bf h}^b(\mathit{goal})$\;
add $x$ to $\mathit{Open}^f$ and $y$ to $\mathit{Open}^b$\;
\While{$\mathit{Open}^f \bigcup \mathit{Open}^b \neq \emptyset$}
{
  {extract from $\mathit{Open}^f \bigcup \mathit{Open}^b$ node $x$ with the smallest $f_1$-value \label{alg:HL:least_cost} \;}
  \nlast
  \lIf{$f_1(x) > \overline{f_1}$} 
  {\textbf{break}}
  
$d \gets$ direction from which $x$ was extracted\;
 $d' \gets$ opposite direction of $d$ \;

      
       \nlast
       $z \gets $ last node in $\mathcal{X}^d(s(x))$ \label{alg:rcebda:quick1}\;
        \nlast
        \lIf{${\bf g}(z) \preceq^\mathrm{Tr} {\bf g}(x)$\label{alg:rcebda:quick2}} 
      {\textbf{continue}}
      
      \nlast
       \lIf{$\mathtt{IsDominated}(x,\mathcal{X}^d(s(x)))$ \label{alg:rcebda:dom}} 
      {\textbf{continue}}

       \nlast
      \ForEach{$y \in \mathcal{X}^d(s(x))$ \label{alg:rcebda:removedom1}}
        { 
        \nlast
        \If {${\bf g}(x) \preceq^\mathrm{Tr} {\bf g}(y)$ }
        {
        \nlast
        move $y$ from $\mathcal{X}^d(s(x))$ to $\mathcal{X}_{\mathit{Dom}}^d(s(x))$ \label{alg:rcebda:removedom3}
        }
        }
        add $x$ to $\mathcal{X}^d(s(x))$ \;
        
    \nlast
    $\mathtt{Match}(x,\mathcal{X}^{d'}(s(x)),\mathcal{X}_{\mathit{Dom}}^{d'}(s(x)))$ \label{alg:rcebda:match}\;

\If{ $g_{\kappa}(x) \leq \overline{f_{\kappa}}/2$} 
    {
    \ForEach{$t \in \mathit{Succ}^d(s(x))$}
        {  $y \gets $ new node with $s(y) = t$ \label{alg:rcebda:expansion1}\; 
             ${\bf g}(y) \gets {\bf g}(x) + {\bf cost}^d (s(x),t)$ \; 
             ${\bf f}(y) \gets {\bf g}(y) + {\bf h}^d (t)$ \;
             $\mathit{parent}(y) \gets x$ \label{alg:rcebda:expansion2}\; 
             
             \nlast
             $z \gets $ last node in $\mathcal{X}^d(t)$\label{alg:rcebda:quick3}\;
             \nlast
             \lIf{ ${\bf f}(y) \npreceq \overline{\bf f}$ \textnormal{\bf or} ${\bf g}(z) \preceq^\mathrm{Tr} {\bf g}(y)$
       \label{alg:rcebda:quick4}} 
      {\textbf{continue}}
      add $y$ to $\mathit{Open}^d$\; \label{alg:rcebda:expansion3}
            
        }
    }

    }
\Return{$\mathit{Sols}$}
\end{algorithm}

%% file: Algs/Match.tex
\begin{algorithm}[t]
\small
\caption{$\mathtt{Match}$}
\label{alg:match}
\DontPrintSemicolon
\KwIn{A node $x$, and sets ($\chi$, $\chi_\mathit{Dom}$) corresponding to $s(x)$}
 
 \KwOutput{$\mathit{Sols}$ updated}

\ForEach{$y \in \chi \bigcup \chi_\mathit{Dom}$}
    {  
    \If {${\bf g}(x) + {\bf g}(y) \preceq {\overline{\bf f}}$\label{alg:match:bound}}
    {
        \If {$g_1(x) + g_1(y) < \overline{f_1}$\label{alg:match:update1}}
        {
        $\overline{f_1} \gets g_1(x) + g_1(y)$ \;
        $\mathit{Sols} \gets \emptyset$\label{alg:match:update2}\;
        }
    $\mathit{dominated} \gets \mathrm{false}$\label{alg:match:dom1}\;
    \ForEach{$(x',y') \in \mathit{Sols}$}
    {
    \If {${\bf g}(x') + {\bf g}(y') \preceq^\mathrm{Tr} {\bf g}(x) + {\bf g}(y)$}
    {$\mathit{dominated} \gets \mathrm{true}$; {\bf break} }
    \If {${\bf g}(x) + {\bf g}(y) \preceq^\mathrm{Tr} {\bf g}(x') + {\bf g}(y')$}
    {remove $(x',y')$ from $\mathit{Sols}$\label{alg:match:dom2}}
    
    }
    \lIf{$\mathit{dominated} = \mathrm{false}$}
    {add $(x,y)$ to $\mathit{Sols}$ \label{alg:match:nondom}}
    }
    
    }

\end{algorithm}

%% file: Figures/Example.tex
\centering
\begin{tikzpicture}[
roundnode/.style={circle, draw=black,  thick, minimum size=5mm},
roundnode2/.style={circle, draw=red,  thick, minimum size=5mm},
scale=0.85, every node/.style={scale=0.90}]

\footnotesize
\node[roundnode,align=center]   at (0, 2) (us)        {$u_s$ \\ (3,3,3)\\ (0,0,0)};
\node[roundnode,align=center]   at (2, 0) (u1)        {$u_1$  \\ (2,3,3) \\ (1,1,1)};

\node[roundnode2,align=center]   at (2, 4) (u2)        {$u_2$  \\ (--,--,3)\\ (--,--,2)};

\node[roundnode,align=center]   at (4, 2) (u3)        {$u_3$ \\ (1,2,2)\\ (2,1,1)};

\node[roundnode,align=center]   at (6, 0) (u4)        {$u_4$ \\ (2,1,1)\\ (3,2,2)};
\node[roundnode2,align=center]   at (6, 4) (u5)        {$u_5$ \\ (--,--,2)\\ (--,--,3)};

\node[roundnode,align=center]   at (8, 2) (ug)        {$u_g$ \\ (0,0,0)\\ (3,3,3)};

\node[roundnode2,align=center,color=gray]   at (0, 0) (u)     {state\\ ${\bf h}^f$\\ ${\bf h}^b$ };

\draw[->,-latex,  thick] (us) edge[auto=right] node[midway, above,rotate=45]{(1,1,2)} (u2);
\draw[->,-latex,  thick] (us) edge[auto=left] node[midway, below,rotate=-45]{(1,1,1)} (u1);
\draw[->,-latex,  thick] (us) edge[auto=left] node[midway, above,rotate=0]{(3,1,1)} (u3);
\draw[->,-latex,  thick] (u2) edge[auto=right] node[midway, above,rotate=-45]{(1,1,1)} (u3);
\draw[->,-latex,  thick] (u1) edge[auto=left] node[midway, below,rotate=45]{(1,1,1)} (u3);
\draw[->,-latex,  thick] (u3) edge[auto=right] node[midway, above,rotate=0]{(1,3,3)} (ug);

\draw[->,-latex,  thick] (u3) edge[auto=left,] node[midway, above,rotate=45]{(1,1,2)} (u5);
\draw[->,-latex,  thick] (u3) edge[auto=left] node[midway, below,rotate=-45]{(1,1,1)} (u4);

\draw[->,-latex,  thick] (u4) edge[auto=right] node[midway, below,rotate=45]{(2,1,1)} (ug);
\draw[->,-latex,  thick] (u5) edge[auto=left] node[midway, above,rotate=-45]{(1,2,2)} (ug);

\end{tikzpicture}


%% file: Tables/example.tex
\begin{table*}[t]
    \small
    \centering
    \setlength{\tabcolsep}{3.5pt}
    \begin{tabular}{ c |  l  c c |l c c | c c}

    \hline
    It. & $\mathit{Open}^f$ : [${\bf f}(x), {\bf g}(x), s(x)$] & $\mathcal{X}^f$ & $\mathcal{X}^f_{Dom}$ & $\mathit{Open}^b$ : [${\bf f}(x), {\bf g}(x), s(x)$] & $\mathcal{X}^b$ & $\mathcal{X}^b_{Dom}$ &  $\mathit{Sols}$\\
    \hline
    1 & $^{*}x_1$=[(3,3,3), (0,0,0), $u_{s}$] & $\mathcal{X}^f(u_s)$=$\{x_1\}$ & 
        & \ \ $x_2$=[(3,3,3), (0,0,0), $u_{g}$] &   &  \\
    
    \hline
    2 & $^{*}x_3$=[(3,4,4), (1,1,1), $u_{1}$] & $\mathcal{X}^f(u_1)$=$\{x_3\}$ & 
        & \ \ $x_2$=[(3,3,3), (0,0,0), $u_{g}$] &   &  \\
     & \ \ $x_4$=[(4,3,3), (3,1,1), $u_{3}$] & & 
        & &   &  \\
    \hline
    3 & $^{*}x_5$=[(3,4,4), (2,2,2), $u_{3}$] & $\mathcal{X}^f(u_3)$=$\{x_5\}$ & 
        & \ \ $x_2$=[(3,3,3), (0,0,0), $u_{g}$] &   &  \\
     & \ \ $x_4$=[(4,3,3), (3,1,1), $u_{3}$] & & 
        & &  &   \\
    
    \hline
    4 & \ \ $x_4$=[(4,3,3), (3,1,1), $u_{3}$]  &  & 
        &$^{*}x_2$=[(3,3,3), (0,0,0), $u_{g}$] & $\mathcal{X}^b(u_g)$=$\{x_2\}$  &  \\
     &  \ \ $x_7$=[(5,4,4), (3,3,3), $u_{4}$]& & 
        & &   &  \\

    \hline
    5 & \ \ $x_4$=[(4,3,3), (3,1,1), $u_{3}$]  &  & 
        &$^{*}x_8$=[(3,4,4), (1,3,3), $u_{3}$] & $\mathcal{X}^b(u_3)$=$ \{x_8\}$ &   \\
     & \ \ $x_7$=[(5,4,4), (3,3,3), $u_{4}$]& & 
        & \ \ $x_9$=[(5,3,3), (2,1,1), $u_{4}$] &  &     \\    
    
    \hline
    6 & $^{*}x_4$=[(4,3,3), (3,1,1), $u_{3}$]  & $\mathcal{X}^f(u_3)$=$\{x_4\}$ & $\mathcal{X}_{Dom}^f(u_3)$=$\{x_5\}$
        & \ \ $x_9$=[(5,3,3), (2,1,1), $u_{4}$] & & &  $\{(x_4,x_8)\}$   \\
     & \ \ $x_7$=[(5,4,4), (3,3,3), $u_{4}$]& & 
        &  &  &   \\
\hline
\end{tabular}
    \caption{Trace of $\mathit{Open}^d$ and $\mathit{Sols}$ in the first 6 iterations (It.) of \textsf{RCEBDA*}. The node extracted in each iteration is marked with symbol $^{*}$. We also show changes on the $\mathcal{X}^d$ and $\mathcal{X}^d_{Dom}$ lists of the explored states in separate columns for both directions.}
    \label{tab:example}
\end{table*}

%% file: Appendix.tex
\section{Theoretical Results}
\label{sec:appendix}
This section formally proves the correctness of \textsf{RCEBDA*}.
Unless otherwise stated, we assume that nodes are compared within the same direction only.
In other words, we do not compare a backward node with a forward node.

\noindent \textbf{Lemma 1\ }
For every $u \in S$, let $h^d_i(u)$ represent the $\mathit{cost_i}$-optimal path from $u$ to a $\mathit{target}$ state on the graph of direction $d$.
$h^d_i$ is consistent and admissible \cite{AhmadiSHJ24}.

\noindent \textbf{Proof Sketch}
Assume the contrary that $h^d_i$ is not consistent, then there exists $(u,v) \in E^d$ for which we have $h^d_i(u) > \mathit{cost}^d_i(u,v) + h^d_i(v)$.
However, the existence of such an edge contradicts our assumption that $h^d_i$ represents optimal costs, because the edge $(u,v)$ can be used to further reduce $h^d_i(u)$ through $h^d_i(v)$.
It is also admissible since we always have $h^d_i(\mathit{target})= 0$ due to $\mathit{cost}_i$-values being non-negative.
\hfill $\square$

\noindent \textbf{Lemma 2\ }
Suppose ${\bf h}^f$ and ${\bf h}^b$ are admissible heuristic functions.
State $u \in S$ cannot be part of any feasible solution path if we have $h^f_i(u) + h^b_i(u) \nleq \overline{f_i}$ for any $i \in \{2,\dots,k\}$.

\noindent \textbf{Proof Sketch}
Given the admissibility of $h^f_i$ and $h^b_i$, any $\mathit{start}$-$\mathit{goal}$ path $\pi$ via $u$ will have $\mathit{cost}_i(\pi) \geq h^f_i(u) + h^b_i(u)$, which yields $\mathit{cost}_i(\pi) > \overline{f_i}$.
Thus, $\pi$ cannot be a solution path due to its $\mathit{cost}_i$ being out of bounds.

\noindent \textbf{Lemma 3\ }
%
Suppose {A*} is led by (smallest) $f_1$-values.
Let $(x_1,x_2,...,x_t)$ be the sequence of nodes explored by {A*} in direction $d$. 
Then, if $h^d_1$ is consistent, $i \leq j$ implies $f_1(x_i) \leq f_1(x_j)$ \cite{hart1968formal}.

\noindent \textbf{Corollary 1\ }
Let $(x_1,x_2,...,x_t)$ be the sequence of nodes extracted from $\mathit{Open}^d$. The heuristic function ${\bf h}^d$ is consistent and admissible. Then, under the premises of Lemma~3, $i \leq j$ implies $f_1(x_i) \leq f_1(x_j)$, meaning $f_1$-values of extracted nodes are monotonically non-decreasing.

\noindent \textbf{Lemma 4\ }
Suppose $x_j$ is extracted after $x_i$ and $s(x_i)=s(x_j).$
$x_i$ weakly dominates $x_j$ if $\mathrm{Tr}({\bf g}(x_i)) \preceq \mathrm{Tr}({\bf g}(x_j))$.

\noindent \textbf{Proof Sketch\ }
$x_j$ is extracted after $x_i$, so we have $f_1(x_i) \leq f_1(x_j)$ according to Corollary~1.
This yields $g_1(x_i) \leq g_1(x_j)$ due to $h^d_1(s(x_i)) = h^d_1(s(x_j))$.
The other condition
$\mathrm{Tr}({\bf g}(x_i)) \preceq \mathrm{Tr}({\bf g}(x_j))$ means $g_2(x_i) \leq g_2(x_j), \dots, g_{k}(x_i) \leq g_{k}(x_j)$.
Thus, ${\bf g}(x_j)$ is no smaller than ${\bf g}(x_i)$ in all dimensions.
\hfill $\square$

\noindent \textbf{Lemma 5\ }
Dominated nodes cannot lead to any non-dominated $\mathit{cost}_1$-optimal solution path.

\noindent \textbf{Proof Sketch\ }
We prove this lemma by assuming the contrary, namely by claiming that dominated nodes can lead to a non-dominated $\mathit{cost}_1$-optimal solution path.
Let $x$ and $y$ be two nodes associated with the same state where $y$ is dominated by $x$.
Suppose that $\pi^*$ is a non-dominated $\mathit{cost}_1$-optimal $\mathit{start}$-$\mathit{goal}$ solution path via the dominated node $y$. 
Since $x$ dominates $y$, one can replace the subpath represented by $y$ with that of $x$ on $\pi^*$ to further reduce the costs of the optimum path through $y$ at least in one dimension (either the primary cost or resources).
However, being able to improve the established optimal solution path would either contradict our assumption on the optimality of $\pi^*$, or its non-dominance in the presence of lower resources.
Therefore, we conclude that dominated nodes cannot form any non-dominated $\mathit{cost}_1$-optimal solution path. 
 \hfill $\square$

  \noindent \textbf{Lemma 6\ }
Node $x$ generated in direction $d$ cannot lead to any $\mathit{cost}_1$-optimal or feasible $\mathit{start}$-$\mathit{goal}$ path if $\mathit{\bf f}(x) \npreceq \overline{\bf f}$.

\noindent \textbf{Proof Sketch\ }
$\mathit{\bf f}(x) \npreceq \overline{\bf f}$ means either $f_1(x) > \overline{f_1}$ or $f_{i}(x) > R_{i-1}$ for a resource $i\in \{2, \dots, k\}$.
The former guarantees that the $x$'s expansion cannot lead to a solution with $\mathit{cost}_1$-value better than any solution in $\mathit{Sols}$ with the optimal cost of $\overline{f_1}$.
Since ${\bf h}^d$ is admissible, the latter condition also ensures that expansion of $x$ does not lead to a solution path within the resource limits.
\hfill $\square$

\noindent \textbf{Lemma 7\ }
Let $y$ be a node weakly dominated by $x$ and $s(x)=s(y)$.
If $y$'s expansion leads to a non-dominated $\mathit{cost}_1$-optimal solution path, $x$'s expansion will also lead to a non-dominated $\mathit{cost}_1$-optimal solution.

\noindent \textbf{Proof Sketch\ }
We prove this lemma by assuming the contrary, namely that $x$ cannot lead to any non-dominated $\mathit{cost}_1$-optimal solution path.
Since $y$ is weakly dominated by $x$, we have $g_i(x) < g_i(y)$ in (at least) one dimension $i\in \{1, \dots, k\}$, or ${\bf g} (x) = {\bf g} (y)$.
The former case means $y$ is dominated by $x$ and thus cannot lead a non-dominated optimal solution (see Lemma~5), contradicting our assumption.
The latter case means both paths are equal in terms of $\mathit{\bf cost}$, thus one can replace the partial path represented by $y$ with that of $x$ to obtain a non-dominated optimal solution path, contradicting our assumption.
Therefore, node $x$ will definitely lead to a non-dominated $\mathit{cost}_1$-optimal solution path if node $y$ also leads to such path.
 \hfill $\square$

\noindent \textbf{Theorem 1\ }
\textsf{RCEBDA*} computes a set of resource-unique non-dominated $\mathit{cost}_1$-optimal paths for the RCSP problem.

\noindent \textbf{Proof Sketch\ }
\textsf{RCEBDA*} enumerates all but promising paths bidirectionally in best-first order (Corollary~1).
The pruning rules ensure that: i) states or nodes violating the upper bounds can be pruned safely (Lemmas~2,6); ii) removal of weakly dominated nodes is safe, as they will not lead to a cost-unique non-dominated optimal solution path (Lemmas~5,7).
Therefore, we just need to show that the algorithm is able to build $\mathit{cost}_1$-optimal paths via the $\mathtt{Match}$ procedure.

Without loss of generality, suppose that $x$ with state $s(x)$ is part of an optimal solution path, and was not expanded in direction~$d$ due to $g_\kappa(x) > \overline{f_\kappa}/2$.
Let $y$ be a node in direction $d'$ that complements $x$, leading to a complete path with ${\bf g}(x) + {\bf g}(y) \preceq \overline{\bf f}$.
Since the joined path is within the bounds, we must have $g_\kappa(y) \leq \overline{f_\kappa}/2$, which ensures the expansion of $y$ in direction $d'$.
We distinguish two cases for $y$:\\
i) $y \in \mathcal{X}^{d'}(s(x)) \bigcup \mathcal{X}^{d'}_{Dom}(s(x))$: we can build the solution path by joining $x$ with $y$ though the $\mathtt{Match}$ procedure.\\
ii) $y \notin \mathcal{X}^{d'}(s(x)) \bigcup \mathcal{X}^{d'}_{Dom}(s(x))$: $y$ has not been explored yet but $x$ is stored in $\mathcal{X}^{d}(s(x))$. Consequently, $x$ can still be joined with $y$ via the $\mathtt{Match}$ procedure after $y$ is extracted.\\
Therefore, the solution path via $x$ and $y$ is always discoverable.
In addition, the explicit dominance check of lines~\ref{alg:match:dom1}-\ref{alg:match:dom2} in Algorithm~\ref{alg:match} guarantees that solutions in $\mathit{Sols}$ remain cost-unique and non-dominated.
Finally, when the search surpasses $\overline{f_1}$, according to Corollary~1, it proves that unexplored nodes are all out of bounds, and consequently, the termination criterion is correct. 
Therefore, we conclude that \textsf{RCEBDA*} terminates with returning a set of resource-unique non-dominated $\mathit{cost}_1$-optimal solution paths.
 \hfill $\square$

%% file: Tables/Main_results.tex
\begin{table}[t]
\centering
\small
\setlength{\tabcolsep}{4.5pt}
\begin{tabular}{|l | l |r | *{3}{r} | *{1}{r}|}
\hline
     Map & Algorithm & $|\mathcal{S}|$ & \multicolumn{1}{c}{t\textsubscript{min}} & \multicolumn{1}{c}{t\textsubscript{avg}} & \multicolumn{1}{c|}{t\textsubscript{max}} & \multicolumn{1}{c|}{$\phi$}\\
  
    \hline
\multicolumn{7}{|c|}{First scenario with two resources ($k=3$)} \\

NY  & \textsf{RCBDA*} & 98  & 0.37 & 153.76 & 3600   & 19.4      \\
    & \textsf{ERCA*}  & 97  & 1.46 & 162.10 & 3600   & 40.0  \\
    & \textsf{RCEBDA*}  & 100  & 0.20 & 1.82 & 29.8   & 1.2  \\
    & \textsf{RCEBDA*\textsubscript{par}}   & 100  & 0.21 & 1.74 & 30.2   & 1.0 \\
\hline
BAY & \textsf{RCBDA*} & 100  & 0.47 & 67.82 & 1434.3   & 8.8      \\
    & \textsf{ERCA*}  & 93  & 1.62 & 290.02 & 3600   & 43.0  \\
    & \textsf{RCEBDA*}  & 100  & 0.24 & 3.12 & 51.3   & 1.1  \\
    & \textsf{RCEBDA*\textsubscript{par}}   & 100  & 0.24 & 2.43 & 33.0   & 1.0 \\
\hline
COL & \textsf{RCBDA*} & 90  & 0.63 & 541.77 & 3600   & 196.1      \\
    & \textsf{ERCA*}  & 79  & 2.23 & 818.82 & 3600   & 95.0 \\
    & \textsf{RCEBDA*}  & 100  & 0.33 & 15.57 & 247.9   & 1.1  \\
    & \textsf{RCEBDA*\textsubscript{par}}   & 100  & 0.33 & 13.45 & 191.8 & 1.0 \\

\hline
\hline

\multicolumn{7}{|c|}{Second scenario with three resources ($k=4$)} \\
NY  & \textsf{RCBDA*} & 94  & 0.50 &  368.90 & 3600 & 16.6\\
    & \textsf{ERCA*}  & 88  & 2.07 & 493.89 & 3600 & 51.4\\
    & \textsf{RCEBDA*}  & 100  & 0.24 & 7.30 & 139.5 & 1.1 \\
    & \textsf{RCEBDA*\textsubscript{par}}   & 100  & 0.24 & 7.71 & 156.7 & 1.0 \\
\hline
BAY & \textsf{RCBDA*} & 94  & 0.63 & 335.70 &  3600 & 14.6\\
    & \textsf{ERCA*}  & 86  & 2.38 & 557.47 & 3600  & 111.6\\
    & \textsf{RCEBDA*}  & 100  & 0.29 & 11.68 & 190.6 & 1.1\\
    & \textsf{RCEBDA*\textsubscript{par}}   & 100  & 0.27 & 9.55 & 138.4 & 1.0\\
\hline
COL & \textsf{RCBDA*} & 80  & 0.83 &  851.59 & 3600 & 159.4 \\
    & \textsf{ERCA*}  & 73  & 3.22 &  1094.83 & 3600 & 81.9\\
    & \textsf{RCEBDA*}  & 100  & 0.41 & 76.70 & 1252.7 & 1.1\\
    & \textsf{RCEBDA*\textsubscript{par}}  & 100  & 0.40 & 72.35 & 1095.4 & 1.1\\

    \hline
\end{tabular}
\caption{Runtime statistics of the algorithms (in seconds) with two and three resources. $|\mathcal{S}|$ is the number of solved cases (out of 100), and $\phi$ shows the average slowdown factor of mutually solved cases with respect to a virtual best oracle. 
Runtime of unsolved instances is considered to be one hour.
}
\label{table:results}
\end{table}

%% file: Figures/ScatterPlot2.tex

\begin{figure}[t]
    \centering
    \begin{tikzpicture}[scale=0.9]
        \begin{axis}[
            width=\columnwidth,
            height=\columnwidth,
            ylabel={Runtime of \textsf{RCEBDA*}/\textsf{RCEBDA*\textsubscript{par}}},
            xlabel={Runtime of \textsf{RCBDA*}},
            xmode=log,
            ymode=log,
            xmin=0.2, xmax=5000,
            ymin=0.2, ymax=5000,
            xticklabels={0.1,1,10,100,1000},
            yticklabels={0.1,1,10,10\textsuperscript{2},10\textsuperscript{3}},
            log basis x={10},
            log basis y={10},
            legend pos=north west,
            legend cell align=left,
            grid=major,
            axis on top,
            legend style={draw=none},
        ]
        
        \addplot[
            only marks,
            mark=o,
            red,
            mark size=2pt,
            line width=0.8pt,
            fill opacity=0.7,
        ]
        table[x=RCBDA, y=RCEBDA] {Figures/ScatterPlot2.dat};
        \addlegendentry{\textsf{RCEBDA*} vs. \textsf{RCBDA*}}

        \addplot[
            only marks,
            mark=x,
            blue,
            mark size=2pt,
            line width=0.8pt,
            fill opacity=0.7,
        ]
        table[x=RCBDA, y=RCEBDA_par] {Figures/ScatterPlot2.dat};
        \addlegendentry{\textsf{RCEBDA*\textsubscript{par}} vs. \textsf{RCBDA*}}

        \addplot[domain=0.05:6000, samples=100, dashed] {x} 
            node[pos=0.55, above, sloped] {$1\times$};
        
        \addplot[domain=0.05:6000, samples=100, dashed] {0.1*x} 
            node[pos=0.65, above, sloped] {$10\times$};

        \addplot[domain=0.05:6000, samples=100, dashed] {0.01*x} 
            node[pos=0.75, below, sloped] {$100\times$};
        
        \end{axis}
    \end{tikzpicture}
    \caption{Runtime distribution of \textsf{RCEBDA*\textsubscript{par}} and \textsf{RCEBDA*} versus \textsf{RCBDA*} over all instances with $k=4$.}
    \label{fig:scatter_plot}
\end{figure}
